\documentclass{article}
\usepackage{ijcai21}

\usepackage{times}
\usepackage{soul}
\usepackage{url}
\usepackage[hidelinks]{hyperref}
\usepackage[utf8]{inputenc}
\usepackage[small]{caption}
\usepackage{graphicx}
\usepackage{amsmath}
\usepackage{amsthm}
\usepackage{booktabs}
\usepackage{algorithm}
\usepackage{algorithmic}
\usepackage{multirow}
\usepackage{amssymb}
\usepackage{nicefrac}
\usepackage{subcaption}
\usepackage{makecell}
\usepackage{enumitem}
\setlist[itemize]{noitemsep, topsep=0pt}

\newcommand{\Bx}{\boldsymbol{x}}
\newcommand{\Bxp}{\boldsymbol{x'}}
\newcommand{\Xadv}{\Bx^{\mathrm{adv}}}
\newcommand{\Lp}{\ell_p}

\title{Demiguise Attack: Crafting Invisible Semantic Adversarial Perturbations \\with Perceptual Similarity}

\author{
Yajie Wang\thanks{Equal contribution.}$^{,1}$\and
Shangbo Wu$^{*,1}$\and
Wenyi Jiang$^1$\and
Shengang Hao$^{1,3}$\and
Yu-an Tan$^{2}$\\ \And
Quanxin Zhang\thanks{Corresponding author.}$^{,1}$
\affiliations
$^1$School of Computer Science and Technology, Beijing Institute of Technology\\
$^2$School of Cyberspace Science and Technology, Beijing Institute of Technology\\
$^3$School of Computer Science and Technology, Nanyang Normal University
\emails
\{wangyajie19, shangbo.wu, jiangwenyi2000, haoshengang, tan2008, zhangqx\}@bit.edu.cn
}

\begin{document}

\maketitle

\begin{abstract}

Deep neural networks (DNNs) have been found to be vulnerable to adversarial examples. Adversarial examples are malicious images with visually imperceptible perturbations. While these carefully crafted perturbations restricted with tight $\Lp$ norm bounds are small, they are still easily perceivable by humans. These perturbations also have limited success rates when attacking black-box models or models with defenses like noise reduction filters. To solve these problems, we propose Demiguise Attack, crafting ``unrestricted'' perturbations with Perceptual Similarity. Specifically, we can create powerful and photorealistic adversarial examples by manipulating semantic information based on Perceptual Similarity. Adversarial examples we generate are friendly to the human visual system (HVS), although the perturbations are of large magnitudes. We extend widely-used attacks with our approach, enhancing adversarial effectiveness impressively while contributing to imperceptibility. Extensive experiments show that the proposed method not only outperforms various state-of-the-art attacks in terms of fooling rate, transferability, and robustness against defenses but can also improve attacks effectively. In addition, we also notice that our implementation can simulate illumination and contrast changes that occur in real-world scenarios, which will contribute to exposing the blind spots of DNNs.

\end{abstract}

\section{Introduction}%
\label{sec:introduction}

Precisely crafted perturbations added onto input data can easily fool DNNs~\cite{Szegedy2014IntriguingPO,Goodfellow2015ExplainingAH,carlini2017towards,Kurakin2017AdversarialEI}. This vulnerability that inherently exists inside DNNs is often exploited with adversarial examples, which are maliciously modified samples with imperceptible perturbations. Adversarial perturbations are crafted in a sense that they should be constrained within a bound that is \textit{as tight as possible}~\cite{Goodfellow2015ExplainingAH,Kurakin2017AdversarialEI,Dong2018BoostingAA}, so that the perturbation is invisible to humans. In light of this, a majority of attacks often measure image similarity with $\Lp$ norms --- $\ell_0$~\cite{carlini2017towards,Papernot2016TheLO}, $\ell_2$~\cite{carlini2017towards} and $\ell_\infty$~\cite{Kurakin2017AdversarialEI,Dong2018BoostingAA}. However, we argue that $\Lp$ norms have limitations.

\begin{figure}[tpb]
  \centering
  \includegraphics[width=\linewidth]{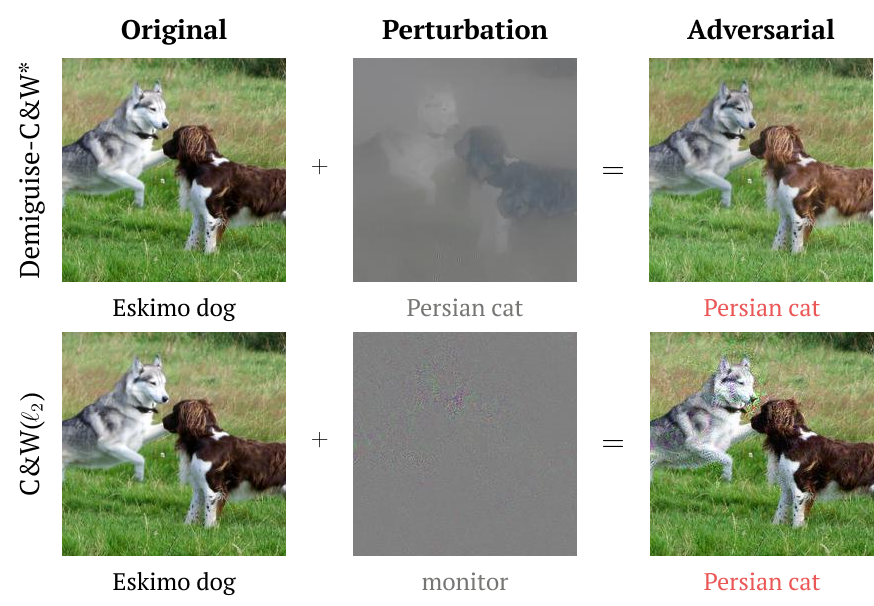}
  \caption{Demonstration of adversarial examples and perturbations crafted by Demiguise-C\&W and C\&W ($\ell_2$) against ResNet-50. C\&W ($\ell_2$) crafts spottable perturbations with arbitrary noise. Demiguise-C\&W crafts much larger perturbations with rich semantic information while maintaining imperceptibility.}
  \label{fig:demiguise-cw-demo}
\end{figure}

Classic per-pixel measurements, such as the $\ell_2$ norm distance, are insufficient for assessing structured data included in images because they assume pixel-wise independence. As such, $\Lp$-based perturbations in the RGB color space often have high spatial frequencies, which inevitably changes the spatial frequencies of natural images, making adversarial examples easily spottable by humans. In addition, the adversary would often need to generate larger perturbations in order to achieve higher adversarial strength. This naturally leads to more perceptible changes to the original image that are, again, noticeable by humans. This is the trade-off between adversarial effectiveness and perturbation imperceptibility, which will always exist if $\Lp$ norm bounds are applied.

Many works on adversarial examples have also gained awareness that pixel-wise differences measured with $\Lp$ norms coincide deficiently with human perception. Many of the non-$\Lp$ solutions execute an attack by modifying the colors of the original image and alleviate the unrealistic factors of the changes with various mitigations, for instance: Semantic Adversarial Examples~\cite{Hosseini2018SemanticAE} and ColorFool~\cite{Shamsabadi2020ColorFoolSA}. But for most of the time, these practices are not enough because many of the color changes are still easily perceptible. An optimal unrestricted adversarial attack should be able to exhibit powerful adversarial effectiveness on the premise that the perturbations are perceptually invisible. However, this unique challenge of comparing image similarity is still a wide-open problem. Because not only are visual patterns highly dimensional and highly correlated themselves, but the very measurement of visual similarity is often quite subjective if it's aimed to mimic human judgments.

In this paper, we propose to attenuate this conundrum by using Perceptual Similarity as a measurement for image similarity when crafting adversarial examples. Perceptual Similarity~\cite{zhang2018unreasonable} is an emergent property shared across deep visual representations. It is a metric that utilizes deep features from trained CNNs to measure image similarity~\cite{Johnson2016PerceptualLF}. Specifically, we optimize perturbations with respect to Perceptual Similarity, creating a novel adversarial attack strategy, namely \textbf{Demiguise Attack}. By manipulating semantic information with Perceptual Similarity, Demiguise Attack can perturb images in a way that correlates extraordinary well with human perception such that the perturbations are imperceptible although they are of large magnitudes, as shown in the middle column of Figure~\ref{fig:demiguise-cw-demo}. Our perturbations with high-order semantic information are even more likely to be classified into the same labels as original images or adversarial examples, which we further demonstrate in Section~\ref{sub:comparison_metrics_perturb_imperceptible}. Larger perturbations make the adversarial examples we craft more powerful, robust, and can even transfer from one task to another, which we discuss in Section~\ref{sub:cross_task_transferability}. We also demonstrate that the effect of our approach is additive and can be used in combination with existing attacks to improve performances further. What's more, as shown in the rightmost column of Figure~\ref{fig:demiguise-cw-demo}, by using Perceptual Similarity, our adversarial examples are somewhat able to simulate the illumination changes that occur in natural situations. This could be used for exposing the blind spots inside the target model and thereby helping it improve under real-world scenarios.

Our key contributions in this paper are:

\begin{itemize}
  \item We propose a novel, unrestricted, black-box adversarial attack based on Perceptual Similarity, called Demiguise Attack. Our approach manipulates semantic information with the HVS-oriented image metric to craft invisible semantic adversarial perturbation.
  \item The perturbations generated with Perceptual Similarity can simulate the illumination and contrast changes in the real-world and enrich the semantic information lying within original images. This phenomenon implies enhancements that potentially exist for DNNs.
  \item Extensive experiments show that Demiguise Attack crafted adversarial perturbations both manifest excellent visual quality and boost adversarial strength and robustness when combined with existing attacks. We demonstrate Demiguise Attack's compelling maximum of 50\% increase in terms of black-box transferability and the promising nearly 90\% successful attacks under cross-task black-box scenarios.
\end{itemize}

\section{Related Work}

\subsection{Adversarial Attack}%
\label{sec:adversarial-attack}


\paragraph{$\Lp$ norm-based adversarial attack.}

Most adversarial attacks utilize a form of $\Lp$ norm-based distance metric, whether they are used as optimization objectives (such as L-BFGS~\cite{Szegedy2014IntriguingPO} and C\&W~\cite{carlini2017towards}), or direct constraints (such as FGSM~\cite{Goodfellow2015ExplainingAH}, BIM~\cite{Kurakin2017AdversarialEI}, and MI-FGSM~\cite{Dong2018BoostingAA}). These adversarial attacks, while being very effective in terms of fooling rate, often generate perturbations that have distinct characteristics --- arbitrary multi-colored noise that is very obvious to human perception.

\paragraph{Non $\Lp$ norm-based adversarial attack.}

Recent work suggests a different approach for adversarial attacks: unrestricted, semantic adversarial examples that are non-$\Lp$ norm-based. Semantic Adversarial Examples~\cite{Hosseini2018SemanticAE} changes the colors of the image by shifting its hues and saturations in the HSV color space. ColorFool~\cite{Shamsabadi2020ColorFoolSA} perturbs images within a chosen natural-color range for specific semantic categories. While these attacks are all non-$\Lp$ based, they all fail to generate perturbations that align well with human perception, creating unrealistic adversarial examples easily noticeable by humans.

\subsection{Perceptual Distance of Images}%
\label{sub:comp_img_sim}

Measuring how similar are two images is often a subjective notion. Per-pixel measurements like the $\ell_2$ Euclidean distance and Peak Signal-to-Noise Ratio (PSNR) have been proven to lack structural representation and fail to account for the many aspects of human perception. Perceptually motivated distance metrics like SSIM~\cite{Wang2004ImageQA} and FSIM~\cite{Zhang2011FSIMAF} are still simple static functions that aren't ideal. Perceptual Similarity~\cite{zhang2018unreasonable}, which utilizes deep features from trained CNNs, models low-level perceptual judgments surprisingly well, outperforming previous widely-used metrics, as we demonstrate in the following section.

\section{Methodology}

\subsection{Measuring Perceptual Similarity}

The challenge of image similarity comparison has been a long-standing problem in the field of computer vision. It also has proven itself to be of significance for generating imperceptible adversarial perturbations. As stated in Section~\ref{sub:comp_img_sim}, classical measurements like $\Lp$-norm distances, PSNR and SSIM, are all inadequate to express the similarity of high-dimensional structured data like images that possess rich information. It would be ideal for constructing a ``distance metric'' that can represent human judgments when measuring image similarity. However, implementing such a metric is challenging, as human judgments are context-dependent, rely on high-order image semantic information, and may not constitute a distance metric. As such, we propose to utilize Perceptual Similarity~\cite{zhang2018unreasonable}, a novel, HVS-oriented metric, to better craft adversarial examples by manipulating the semantic information lying within images. Here we briefly address concepts of Perceptual Similarity under adversarial settings.

Perceptual Similarity is a novel image quality metric that extracts characteristics from deep feature spaces of CNNs trained on image classification tasks. The metric itself is neither a special function nor a static module; instead, it is a consequence of visual representations tuned to be predictive about real-world structured information. Hence, we would be more capable of cultivating rich semantic information for crafting adversarial perturbation if we were to utilize Perceptual Similarity, thereby calibrating our perturbation to be in line with human perception. Notably, Perceptual Similarity is calculated based on a predefined and pretrained perceptual similarity network. For an original image $\Bx$ and its distorted partner $\Bxp$, with perceptual similarity network $\mathcal{N}$, we compute the distance between $\Bx$ and $\Bxp$ as
\begin{equation}
  \label{eq:lpips-distance}
  \mathcal{D}(\Bx,\Bxp) = \sum_l \frac{1}{H_l W_l} \cdot \sum_{h,w}\|\omega_l \odot
  (\hat{\theta}^l_{hw} - \hat{\theta'}^l_{hw})\|^2.
\end{equation}
where $l$ is one of the layers in the $L$ layers feature stack in network $\mathcal{N}$, $\hat{\theta}^l, \hat{\theta'}^l \in \mathbb{R}^{H_l\times W_l\times C_l}$ are normalized features extracted from $\Bx$ and $\Bxp$ when they are passing through layer $l$ (with $H_l,W_l,C_l$ representing height, width, and channel for each layer $l$ respectively), and $\omega_l$ is the vector that is used to scale the channel-wise activations. For adversarial attacks, $\Bx$ is the input image, and $\Bxp$ is the resulting adversarial example. In this setting, the adversary attempts to make the classifier mispredict by modifying $\Bx$ with a negligible perturbation, producing $\Bxp$ that is as close to $\Bx$ as possible --- the optimization goal. The \textit{closeness} between $\Bx$ and $\Bxp$ is, in our case, measured by Eq.~\ref{eq:lpips-distance}. Hence, Eq.~\ref{eq:lpips-distance} also constitutes as a perceptual loss function. By optimizing against Perceptual Similarity, we can manipulate deep semantic features that exist within natural images to create perturbations that correlate well with human perception.

\subsection{Demiguise Attack}

\begin{figure}[tbp]
  \centering
  \includegraphics[width=1.0\linewidth]{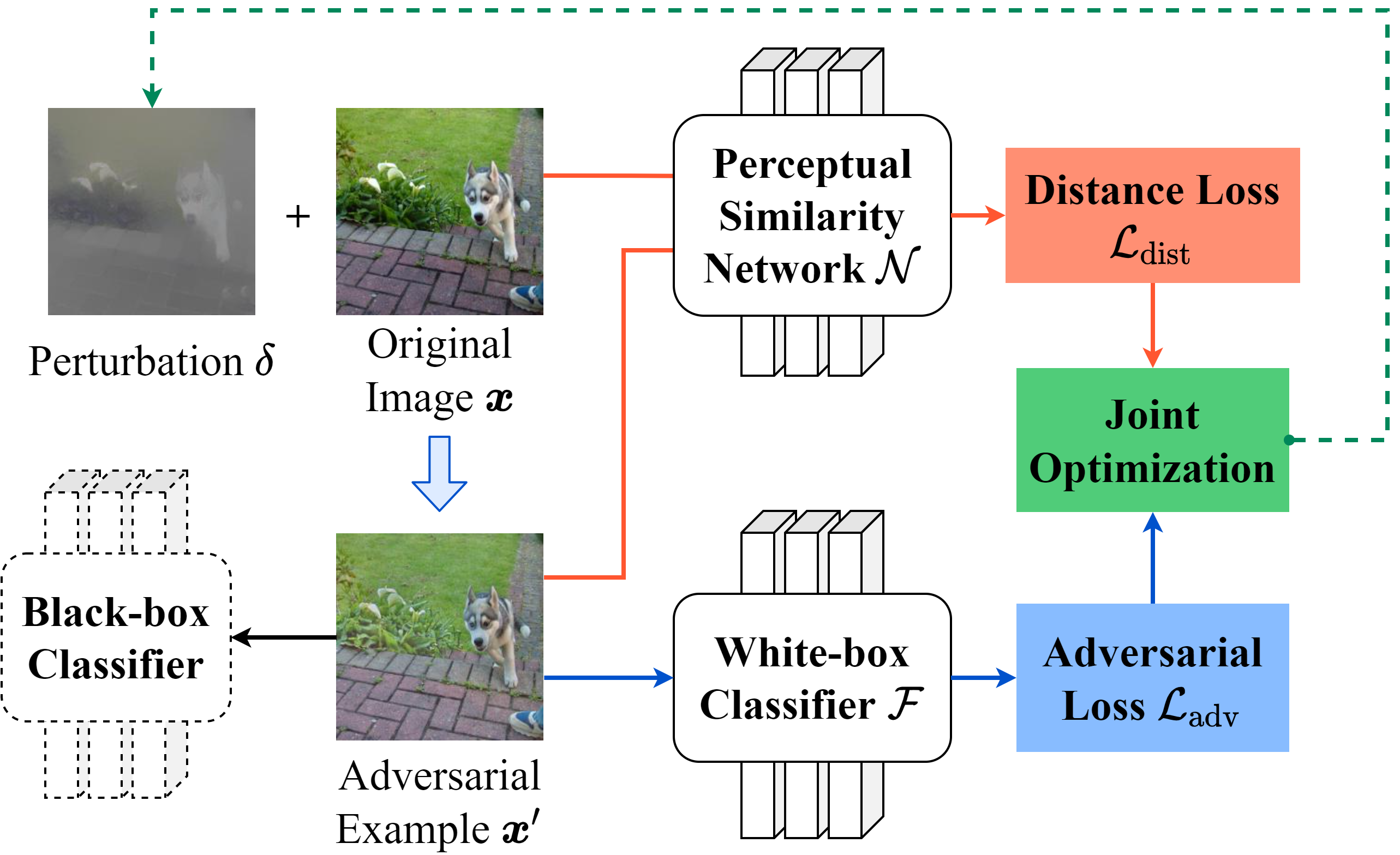}
  \caption{General optimization procedure of Demiguise Attack. For each iteration, we obtain both the $\mathcal{L}_\mathrm{dist}$ from Perceptual Similarity network $\mathcal{N}$ as a penalty, and the $\mathcal{L}_\mathrm{adv}$ from the classifier. Both of them are then optimized as $\mathcal{L}=\lambda\cdot\mathcal{L}_\mathrm{dist}+\mathcal{L}_\mathrm{adv}$ directly in the back-propagation procedure for crafting adversarial perturbations.}
  \label{fig:figs/demiguise-extend-method}
\end{figure}

\paragraph{Demiguise-C\&W --- Combining C\&W with Demiguise Attack} Demiguise Attack is a universal strategy for integrating Perceptual Similarity into adversarial attacks. We start with Demiguise-C\&W, which, as its name suggests, is a variant of Demiguise Attack with its optimisation solved using techniques from C\&W~\cite{carlini2017towards}. Classic adversarial attacks often use optimization procedures to craft perturbations. In order to satisfy the adversary's goal, one would optimize against an objective function in order to find minimal perturbations as
\begin{equation}
  \Xadv=\underset{\Bxp:\mathcal{F}(\Bxp)\neq y}{\arg\min} \|\Bxp-\Bx\|_p.
\end{equation}
for non-targeted attacks, where $\Bx$ and $\Bxp$ are the original image and its (intermediate) adversarial example respectively, $\mathcal{F}$ is the target classifier, $p$ is the norm used for distance calculation, and $\Bx^{\mathrm{adv}}$ is the final adversarial example. This is basically an optimization against a loss function that perturbs the input image until it is adversarial, while also assuring that the perturbation is minimal. Formally, current adversarial attacks are optimizations against the joint loss of
\begin{equation}
  \mathcal{L} = \mathcal{L}_{\mathrm{adv}} + \mathcal{L}_{\mathrm{dist}}.
\end{equation}
in essence, where $\mathcal{L}_{\mathrm{adv}}$ is the adversarial loss that guides the optimization procedure to making the adversarial example \textit{adversarial}, and $\mathcal{L}_{\mathrm{dist}}$ is the distance loss that minimizes the distance between the original input and the adversarial example. However, this optimization problem is often thought to be NP-hard, and as such, various mitigation measures have been proposed to solve it. C\&W attack, as one of the most effective approaches, use an $f$ function-interpreted cross-entropy loss as $\mathcal{L}_{\mathrm{adv}}$
\begin{equation}
  f(\Bxp)=\max(\max\{Z(\Bxp)_i:i\neq t\}-Z(\Bxp)_t,0).
  \label{eq:f_function}
\end{equation}
where $Z(\Bxp)$ is the logit of $\Bxp$. Demiguise Attack applies optimizations against Perceptual Similarity with similar approaches as in C\&W, formally expressed as
\begin{equation}
  \underset{\boldsymbol{u}}{\mathrm{minimize}}\ \lambda\cdot \mathcal{D}(\Bx,\Bxp)+f(\Bxp).
\end{equation}
where a change of variables is applied, making $\Bxp = \nicefrac{1}{2}\cdot\tanh{(\boldsymbol{u})} + 1$, so that we optimise over $\boldsymbol{u}$ instead of $\Bxp$ directly. We take the five convolutional layers from the VGG architecture as the perceptual similarity network $\mathcal{N}$ with pretrained weights. The perceptual similarity network $\mathcal{N}$ exposes distance $\mathcal{D}(\boldsymbol{x}, \boldsymbol{x'})$ along with gradient information $\boldsymbol{g}$ so we can fully utilize it for crafting adversarial perturbations. The general Demiguise-C\&W strategy is expressed in detail in Algorithm~\ref{alg:general_algo_of_demi_attack}.

\begin{algorithm}[tbp]
\caption{Demiguise-C\&W}%
\label{alg:general_algo_of_demi_attack}
\begin{algorithmic}[1]
  \REQUIRE{Input image $\Bx$, original prediction $y$, weight of Perceptual Similarity loss $\lambda$, number of iterations $N$;}
  \ENSURE{Adversarial example $\Bxp$;}
  \STATE{Initialize: $\boldsymbol{x_0'} \leftarrow \Bx$, $\boldsymbol{u}_0
  \leftarrow 0$;}
  \STATE{Construct $\mathcal{D}$ --- Perceptual Similarity instance with pretrained network $\mathcal{N}$;}
  \FOR{$i = 0$ \TO$N - 1$}
  \STATE{Initialize perturbation:
    $\boldsymbol{\delta}_i\leftarrow\nicefrac{1}{2}\cdot\tanh(\boldsymbol{u}_i) + 1 - \Bx$;}
  \STATE{$\mathcal{L}_{\mathrm{dist}} \leftarrow \mathcal{D}(\boldsymbol{\delta}_i +\Bx, \Bx)$;}
  \STATE{$\mathcal{L}_{\mathrm{adv}}\leftarrow f(\boldsymbol{\delta}_i + \Bx)$. $f$ is specified in Equation~\ref{eq:f_function};}
  \STATE{Minimize $\boldsymbol{u}_i$ over $\mathcal{L} = \lambda\cdot\mathcal{L}_{\mathrm{dist}} + \mathcal{L}_{\mathrm{adv}}$;}
  \STATE{$\boldsymbol{x'_i} \leftarrow \nicefrac{1}{2} \cdot \tanh(\boldsymbol{u}_i) + 1$;}
  \IF{$\mathcal{L}$ is not converging}
    \STATE{Early stop and return $\boldsymbol{x'_i}$;}
  \ENDIF
  \ENDFOR
  \RETURN{$\Bxp \leftarrow \boldsymbol{x'_i}$}
\end{algorithmic}
\end{algorithm}

\paragraph{Combining other attacks with Demiguise Attack} More importantly, we demonstrate that the strategy of Demiguise Attack is additive, and it can be used in combination with existing attacks to improve performances further. We showcase our strategy's simplicity and universality by illustrating how to combine Demiguise Attack with existing attacks.

In Demiguise Attack, Perceptual Similarity is used directly as the distance penalty for optimization-based attacks. Besides the actual distance, we can also access gradient information from the Perceptual Similarity distance as $\boldsymbol{g}=\nabla_{\Bx} \mathcal{D}(\Bx,\Bxp)$, which can then be used in gradient-based attacks. We address our general Demiguise Attack strategy (perturbation optimization procedure) in Figure~\ref{fig:figs/demiguise-extend-method}.

With this, we can use Demiguise Attack in combination with existing state-of-the-art attacks, creating Demiguise-\{C\&W, MI-FGSM, HopSkipJumpAttack\}. Here we briefly address their implementations. For C\&W and HopSkipJumpAttack~\cite{chen2020hopskipjumpattack}, we combine their optimization procedure with Perceptual Similarity distance $\mathcal{D}$. For MI-FGSM, we update its loss as
\begin{equation}
  \mathcal{L}=\nabla_{\Bx}\mathcal{J}(\Bxp, y) + \lambda \cdot \nabla_{\Bx} \mathcal{D}(\Bxp,\Bx)
\end{equation}
where $\mathcal{J}$ is our usual cross-entropy loss. For all Demiguise variants of these attacks, we introduce a new vector $\lambda$ to balance the joint-optimization of perceptual loss and adversarial loss. Using Perceptual Similarity, Demiguise Attack crafts adversarial examples that dig deep into the rich semantic representations of images, achieving superior adversarial effectiveness while maintaining compelling imperceptibility.

\begin{figure}[tbp]
  \centering
  \includegraphics[width=\linewidth]{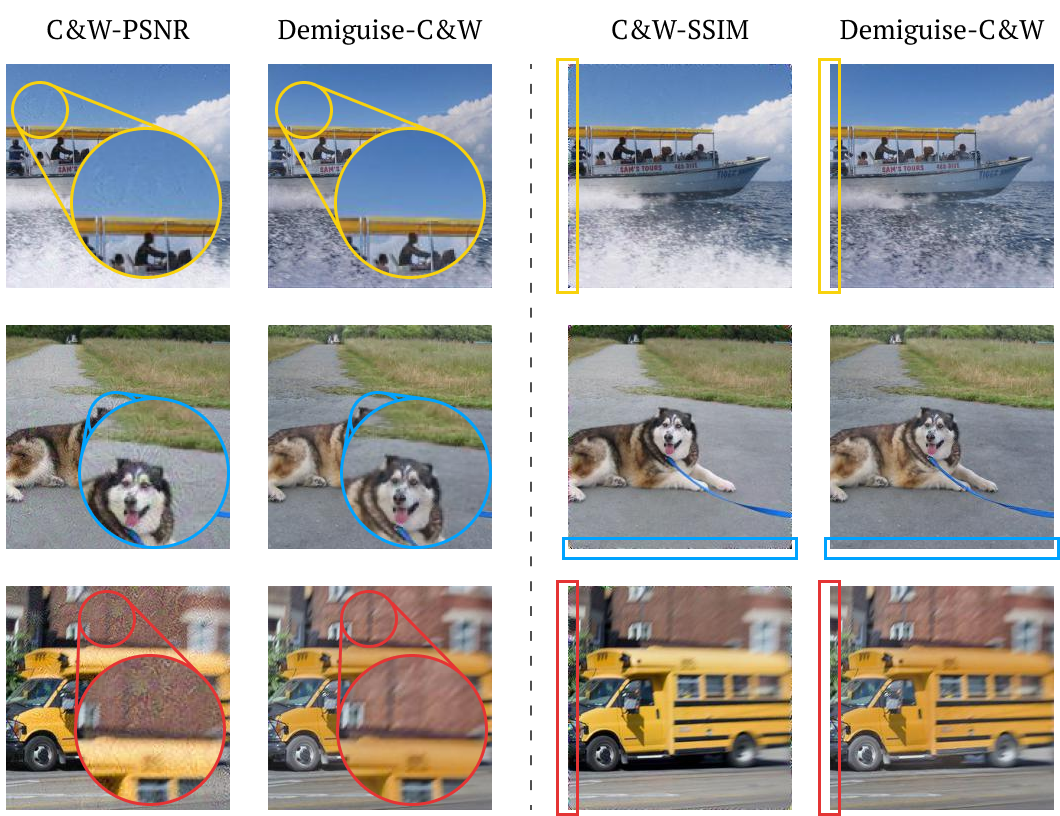}
  \caption{We compare perturbation imperceptibility of our Demiguise-C\&W with C\&W-PSNR and C\&W-SSIM.\@ We find that while all three attacks achieve 100\% fooling rate, only Demiguise-C\&W crafted adversarial examples are able to truly maintain perturbation imperceptibility.}%
  \label{fig:figs/metric-comparison}
\end{figure}

\section{Experiments}

\subsection{Experiment Setup}

We use four different architectures of classifiers for evaluation: ResNet-\{18, 50, 101, 152\}~\cite{He2016DeepRL}, VGG-\{11, 16, 19\}~\cite{Simonyan2015VeryDC}, Inception-v3~\cite{Szegedy2016RethinkingTI}, MobileNet-v2~\cite{Sandler2018MobileNetV2IR}. We use pretrained models from PyTorch's \texttt{torchvision} library. We randomly pick 1000 images of 10 separate classes from ImageNet~\cite{Deng2009ImageNetAL} that are all classified correctly by the models. All images are resized to $256*256$ and center cropped to $224*224$ in size, then normalized with \texttt{mean=[0.485, 0.456, 0.406]} and \texttt{std=[0.229, 0.224, 0.225]}. Our experiments are run on Ubuntu 20.04 LTS with NVIDIA GeForce RTX™ 3090 GPUs and 64GB of memory.

In terms of attacks, we choose a learning rate of $0.2$ and a maximum of 1000 iterations for Demiguise-C\&W. For Demiguise-MI-FGSM, we choose an $\epsilon$ of 0.4, a max iteration of $70$ rounds, and a decay factor of $1.0$. For Demiguise-HSJA, we choose a max iteration of 2000 queries. We use the $\Lp$-norm based versions of these attacks as baselines (including C\&W~\cite{carlini2017towards}, MI-FGSM~\cite{Dong2018BoostingAA}, and HopSkipJumpAttack~\cite{chen2020hopskipjumpattack}), and we use the same hyper-parameters. Adversarial strength is evaluated in terms of the fooling rate of the attack, i.e., the proportion of successful cases over the whole validation set.


\subsection{Comparison of Different Perceptual Distances}%
\label{sub:comparison_metrics_perturb_imperceptible}

\begin{figure*}[tbp]
  \centering
  \includegraphics[width=0.96\linewidth]{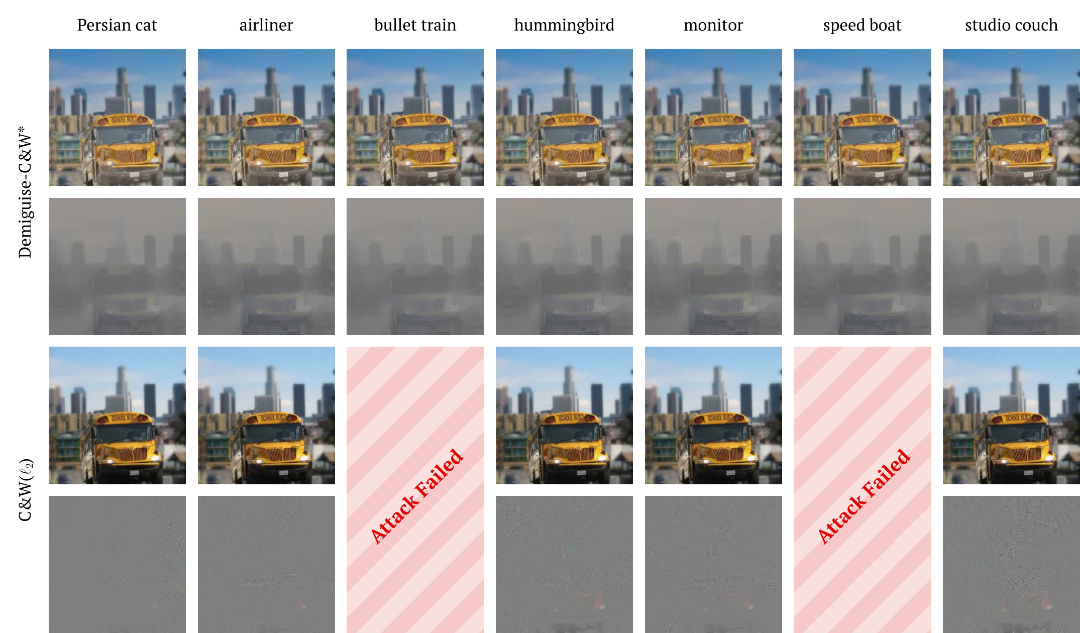}
  \caption{Comparing perturbation crafted by Demiguise-C\&W and C\&W ($\ell_2$) respectively for seven targeted white-box attack scenarios. Demiguise-C\&W both generates perturbation that achieves compelling imperceptibility, and succeeds for all seven attacks.}%
  \label{fig:figs/perturbation_comparison}
\end{figure*}

We start our experiments by investigating whether Perceptual Similarity is the optimal metric for crafting adversarial perturbations with respect to human perception. We compare the aforementioned metrics, including PSNR and SSIM, with Perceptual Similarity. We extend C\&W attack as C\&W-PSNR and C\&W-SSIM and attack the same models as Demiguise-C\&W with the same parameters.

Here we illustrate a few of the adversarial examples crafted by these attacks. Despite the fact that PSNR and SSIM are some of the most commonly used metrics in computer vision, they still are shallow, facile functions that fail to account for the many factors of human perception. We can see that perturbations crafted by C\&W-PSNR and C\&W-SSIM contain high spatial frequencies, which changes the spatial frequencies of natural images, as shown in Figure~\ref{fig:figs/metric-comparison}. C\&W-PSNR crafted examples have obvious arbitrary noise, as shown in the enlarged region. Although C\&W-SSIM's perturbations are less perceptible, a lot of noise is generated along the four sides of the image, as shown in the circled part. These noises make the perturbations spottable by humans and are the major drawbacks of these simple metrics. Conversely, Demiguise-C\&W takes advantage of Perceptual Similarity, perturbing images by probing the rich semantic information within high-order structured representations, generating imperceptible perturbations. Besides, we couldn't help but notice that Demiguise Attacks' perturbations can somewhat simulate the natural illumination changes that occur in real-world situations. We believe that this peculiar phenomenon may suggest potential improvements that exist for current DNNs.

We are intrigued to see whether our perturbations crafted by Demiguise Attack actually mean anything to our ``smart'' classifiers. Thus, we passed our Demiguise Attack's perturbations through the classifiers alone and calculated the ratio of which the ``perturbation'' itself is classified as the same prediction of the original image or the adversarial example. We find that Demiguise Attack's perturbations attain ratios of about 80\% for all four classifiers, more than 40\% higher than those of C\&W ($\ell_2$) as shown in Table~\ref{tab:pert_sem_rate}. Hence, we argue that Demiguise Attack is truly utilizing the high-dimensional semantic information that was not fully used by other attacks. This behavior of semantic information manipulation is what empowers Demiguise Attack to create large feature-rich perturbations that maintain excellent imperceptibility.

\begin{figure}[t]
  \centering
  \begin{subfigure}{\linewidth}
  \begin{center}
    \includegraphics[width=0.97\linewidth]{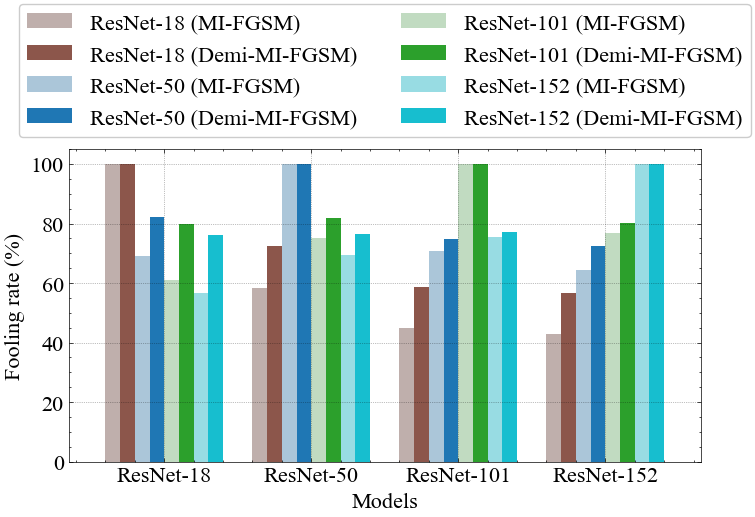}
  \end{center}
  \caption{Transferability among ResNet models}%
  \label{fig:figs/mifgsm_trans_resnet_fam}
  \end{subfigure}\vspace{3mm}
  \begin{subfigure}{\linewidth}
  \begin{center}
    \includegraphics[width=0.97\linewidth]{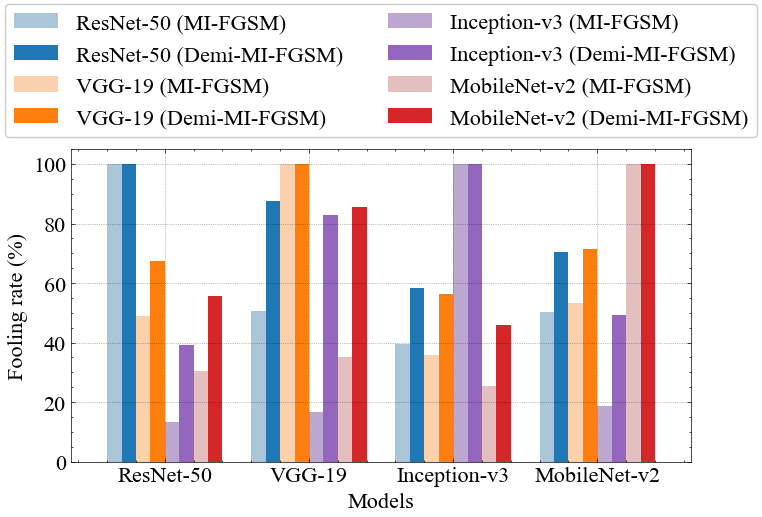}
  \end{center}
  \caption{Transferability among models with different architectures}%
  \label{fig:figs/mifgsm_trans_diff_arch}
  \end{subfigure}
  \caption{We compare the transferability of Demiguise-MI-FGSM with $\ell_\infty$-based MI-FGSM.\@ Fooling rates that reach 100\% are white-box attacks. Apart from these, we find that Demiguise-MI-FGSM (solid colors in the chart) outperforms $\ell_\infty$-based MI-FGSM (translucent colors in the chart) across all models in terms of transferability. Fooling rates can be increased by an average of 10\% to 30\% by incorporating Demiguise Attack strategy.}
  \label{fig:figs/mifgsm_trans_compare}
\end{figure}

We further discuss the intriguing characteristics of our feature-rich perturbations crafted with respect to perceptual similarity. Here we demonstrate a comparison of perturbation between C\&W ($\ell_2$) and our Demiguise-C\&W under targeted white-box scenarios. For the ground truth image that is correctly classified as ``school bus'' by ResNet-50, we attack it for a total of seven times, each targeted to a different class. The results are shown in Figure~\ref{fig:figs/perturbation_comparison}. Once again, we observe that our perturbation crafted by Demiguise-C\&W supersedes those of C\&W ($\ell_2$) in terms of perturbation imperceptibility. Not only did Demiguise-C\&W succeeded in all seven targeted attacks, but it also managed to maintain outstanding adversarial example quality, not to mention that $\ell_2$-based C\&W even failed twice in the seven attacks.

\begin{table}[tpb]
  \centering
  \scalebox{0.8}{%
  \begin{tabular}{lcccc}
    \toprule
    \textbf{Attack} & \textbf{ResNet-50} & \textbf{VGG-19} & \textbf{Inception-v3} & \textbf{MobileNet-v2} \\ \midrule
    Demi-C\&W* & {\bf 81.11\%} & {\bf 86.51\%} & {\bf 72.51\%} & {\bf 81.35\%} \\
    C\&W ($\ell_2)$ & 39.86\% & 48.30\% & 35.72\% & 40.60\% \\
    \bottomrule
  \end{tabular}}
  \caption{We pass our crafted perturbations alone into the same classifiers, and see whether these perturbations contain semantic meanings to our classifiers. Demiguise-C\&W crafted perturbations alone are classified into the same label as the original image or the adversarial example for a ratio of around 80\%, which is over 40\% higher than those crafted by $\ell_2$-based C\&W.}%
  \label{tab:pert_sem_rate}
\end{table}

\subsection{Extend Adversarial Attacks with Demiguise Attack}

With the excellent performances of Demiguise Attack, we continue to combine our approach with other state-of-the-art attacks. First, we combine Demiguise Attack with MI-FGSM and investigate whether our strategy can further improve its black-box transferability. For this experiment, we use MI-FGSM and Demiguise-MI-FGSM to attack (1) the ResNet family (including ResNet-\{18, 50, 101, 152\}), and (2) ResNet-50, VGG-19, Inception-v3, MobileNet-v2. The fooling rates of these attacks under transfer-based black-box scenarios are shown in Figure~\ref{fig:figs/mifgsm_trans_resnet_fam} and Figure~\ref{fig:figs/mifgsm_trans_diff_arch} respectively, where translucent-colored bars represent fooling rates of $\Lp$-based MI-FGSM, and solid-colored bars represent those of Demiguise-MI-FGSM.\@ We find that the performance of MI-FGSM can be increased by an outstanding margin of 10\% to 30\% overall by utilizing our Demiguise Attack strategy despite whether models are of the same or different architectures.

\begin{figure}[t]
  \centering
  \includegraphics[width=0.97\linewidth]{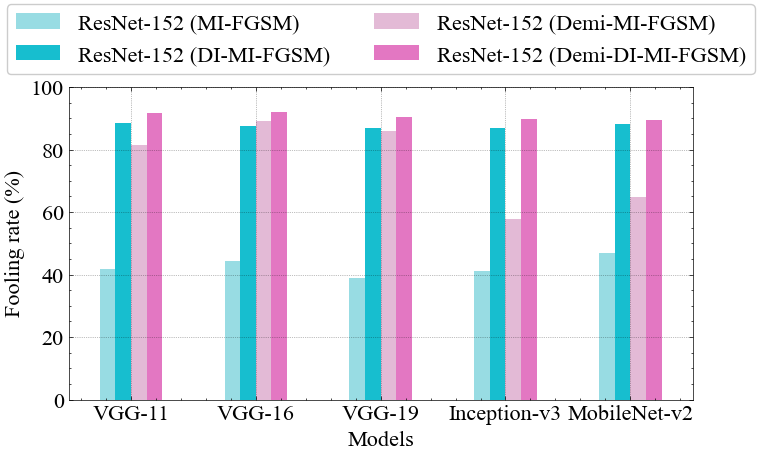}
  \caption{We testify if incorporating input diversity into Demiguise-MI-FGSM ($\ell_\infty$) can further improve transferability. We find that our final attack: Demiguise-DI-MI-FGSM ($\ell_\infty$), surpasses all other attacks' fooling rates, improving DI-MI-FGSM's performance by more than 5\%, reaching a maximum transfer-based fooling rate of over 90\%.}%
  \label{fig:figs/mifgsm_trans_di}
\end{figure}

Input diversity is another approach that has been proven to be of great contribution to the transferability of MI-FGSM.~\cite{Xie2019ImprovingTO} Here, we attack white-box ResNet-152, and transfer generated adversarial examples to VGG-\{11, 16, 19\}, Inception-v3 and MobileNet-v2. We compare the transferability of Demiguise-MI-FGSM, Demiguise-DI-MI-FGSM with baseline performances respectively. As shown in Figure~\ref{fig:figs/mifgsm_trans_di}, we can see that input diversity (DI-MI-FGSM) improves the transferability by a large extent, coming close to or even surpassing the performance of Demiguise-MI-FGSM.\@ By incorporating our approach with input diversity, we can further increase the transferability for more than 5\%, reaching fooling rates as high as 92\%, creating one of the strongest attacks in terms of black-box transferability.

\begin{table}[tpb]
  \centering
  \scalebox{0.8}{%
  \begin{tabular}{lcccc}
    \toprule
    \textbf{Attack} & \textbf{ResNet-50} & \textbf{VGG-19} & \textbf{Inception-v3} & \textbf{MobileNet-v2} \\ \midrule
    ColorFool & 48.1\% & 35.0\% & 40.4\% & 42.9\% \\
    HSJA ($\ell_2$) & 94.9\% & 92.2\% & 69.1\% & 89.7\% \\
    Demi-HSJA* & {\bf 94.9\%} & {\bf 92.0\%} & {\bf 99.1\%} & {\bf 89.4\%} \\
    \bottomrule
  \end{tabular}}
  \caption{We compare the fooling rates of $\ell_2$-based HopSkipJumpAttack, Demiguise-HSJA, and ColorFool under black-box scenarios. The fooling rate of ColorFool barely reaches 50\% on 1000 samples, while Demiguise-HSJA keeps up with HopSkipJumpAttack ($\ell_2$), reaching fooling rates of over 89\%.}%
  \label{tab:adversatial_effectiveness_comp}
\end{table}

\begin{figure}[t]
  \centering
  \includegraphics[width=\linewidth]{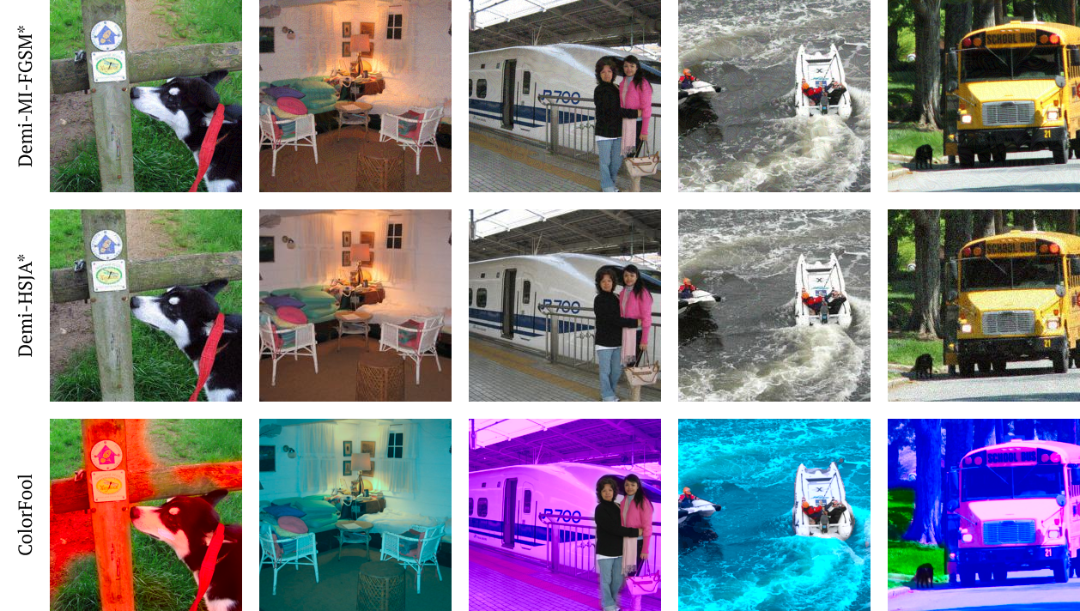}
  \caption{We demonstrate a few of the adversarial examples generated by Demiguise-\{MI-FGSM,HSJA\}, and ColorFool. Although ColorFool lives up to its promise of modifying specific semantic regions, the colors of its final examples are obvious and misleading.}%
  \label{fig:figs/demi-hsja-colorfool}
\end{figure}

Finally, we combine Demiguise Attack with one of the most potent decision-based black-box attacks: HopSkipJumpAttack (HSJA). As previously mentioned, ColorFool~\cite{Shamsabadi2020ColorFoolSA} is a typical non-$\Lp$ based black-box adversarial attack where it only changes the colors of specific semantic categories in an image. We compare these attacks under black-box scenarios. A few of the examples crafted by these attacks are shown in Figure~\ref{fig:figs/demi-hsja-colorfool}. We can easily see that ColorFool's color-changing strategy is not exactly ideal in terms of perturbation imperceptibility, generating colors beyond human comprehension, while Demiguise-HSJA crafted perturbations are imperceptible. What's more, ColorFool barely reaches fooling rates of over 50\% over 1000 samples. In contrast, Demiguise-HSJA, with fooling rates of over 89\%, keeps up with the performance of HopSkipJumpAttack ($\ell_2$). Detailed performances of ColorFool and Demiguise-HSJA are shown in Table~\ref{tab:adversatial_effectiveness_comp}.

\subsection{Adversarial Robustness under Defense}%
\label{sub:adversarial_robustness_under_defense}

In order to gain deeper insight into the effectiveness and strength of Demiguise Attack, we further testify the robustness of our attack against common adversarial defense strategies. Specifically, we utilize both JPEG compression~\cite{Das2018SHIELDFP} and binary filter~\cite{Xu2018FeatureSD} as preprocessing measures that we implement directly into our aforementioned classifiers. We set JPEG compression quality as 75 and binary filter bit-depth as 4.

\renewcommand\theadalign{cc}
\renewcommand\theadfont{\bfseries}
\renewcommand\theadgape{\Gape[2pt]}
\begin{table}[tpb]
  \centering
  \scalebox{0.85}{
  \begin{tabular}{ccccc}
    \toprule
    \thead{Models} & \thead{Attacks} & \thead{No\\Defense} & \thead{JPEG\\
    Compress} & \thead{Binary\\Filter} \\ \midrule
    \multirow{2}{*}{ResNet-50} & C\&W ($\ell_2$) & 100\% & 82.09\% & 81.50\% \\
                               & Demi-C\&W* & 100\% & {\bf 85.10\%} & {\bf
                               84.70\%} \\ \midrule
    \multirow{2}{*}{VGG-19} & C\&W ($\ell_2$) & 100\% & 84.40\% & 81.59\% \\
                            & Demi-C\&W* & 100\% & 84.40\% & {\bf 85.80\%} \\
                            \midrule
    \multirow{2}{*}{Inception-v3} & C\&W ($\ell_2$) & 100\% & 80.40\% & 88.80\% \\
                            & Demi-C\&W* & 100\% & {\bf 83.50\%} & {\bf 92.20\%}
                            \\ \midrule
    \multirow{2}{*}{MobileNet-v2} & C\&W ($\ell_2$) & 100\% & 76.00\% & 84.59\% \\
                            & Demi-C\&W* & 100\% & {\bf 76.70}\% & 79.30\% \\
    \bottomrule
  \end{tabular}}
  \caption{We use JPEG compression and binary filters as defense schemes, and testify the fooling rates of C\&W and Demiguise-C\&W against these defenses.  We find that Demiguise-C\&W achieves 3\% to 5\% better robustness compared with $\Lp$-based C\&W for most of the models with active defenses.}%
  \label{tab:robustness_under_defense}
\end{table}

We demonstrate the fooling rates of C\&W and Demiguise-C\&W in Table~\ref{tab:robustness_under_defense}. Due to the fact that defense schemes like JPEG compression and binary filter try to remove adversarial perturbations by image preprocessing filters, we find that Demiguise-C\&W crafted adversarial perturbations tend to stay more intact than $\Lp$-based ones throughout defenses. We observe a 3\% to 5\% increase in fooling rates for Demiguise-C\&W compared with $\Lp$-based C\&W when attacking models with defenses.

\begin{figure}[t]
  \begin{subfigure}{\linewidth}
  \begin{center}
    \includegraphics[width=0.9\linewidth]{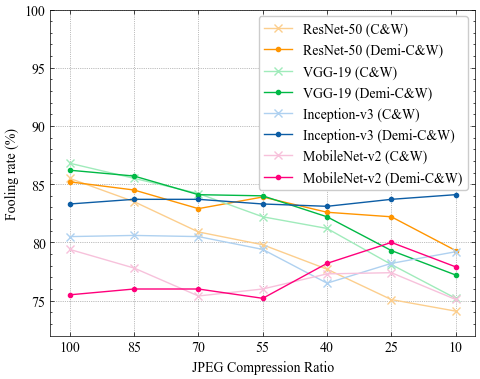}
  \end{center}
  \caption{JPEG Compression}%
  \label{fig:figs/jpeg_compression}
  \end{subfigure}
  \begin{subfigure}{\linewidth}
  \begin{center}
    \includegraphics[width=0.9\linewidth]{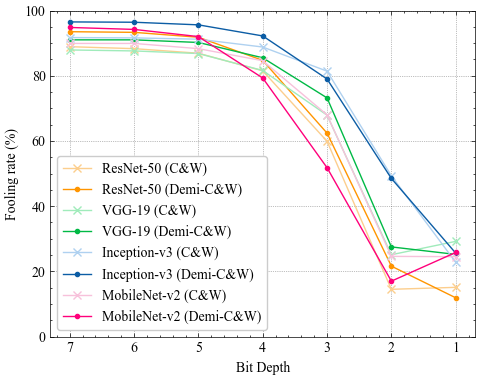}
  \end{center}
  \caption{Binary Filter}%
  \label{fig:figs/binary_filter}
  \end{subfigure}
  \caption{We compare the adversarial robustness of Demiguise-C\&W with C\&W ($\ell_2$) with defenses of different parameters. When image quality is reasonable, we find that no matter what type of defense scheme is applied, Demiguise-C\&W always achieves higher fooling rates than $\ell_2$-based C\&W.}%
  \label{fig:figs/defenses}
\end{figure}

Moreover, both JPEG Compression and Binary Filter have parameters that we can tweak: the compression ratio for JPEG Compression, and the bit depth for Binary Filter. For JPEG compression, we set the compression ratio from 100 to 10 with a step size of 15. For Binary Filter, we set its bit depth from 7 to 1 with a step size of 1. The results of JPEG Compression as a defense scheme are shown in Figure~\ref{fig:figs/jpeg_compression}, and the results of Binary Filter are shown in Figure~\ref{fig:figs/binary_filter}. We can see for both of these defense strategies, the lower image quality (i.e., lower JPEG compression ratio or the smaller bit depth), the more all fooling rates decreases. This characteristic is especially obvious for Binary Filter, where we see clear drops of our fooling rates when bit depth decreases. Nevertheless, when image quality is still reasonable, we find that Demiguise-C\&W still achieves higher fooling rates than vanilla C\&W for all four classification models most of the time. We argue that the utilization of perceptual similarity-based optimization rewards us with this boost in adversarial robustness and effectiveness.

\subsection{Cross-task Transferability}%
\label{sub:cross_task_transferability}

\begin{table}[t]
  \centering
  \scalebox{0.68}{%
  \begin{tabular}{ccccccc}
  \toprule
  \thead{Target\\Classifiers} & \thead{Attacks} & \thead{Black-box\\Detectors} & \thead{Label\\Vanish} &
    \thead{Mis-\\label} & \thead{Extra\\Object} & \thead{Success\\Cases} \\
    \midrule
  \multirow{4}{*}[-2pt]{ResNet-50} &
  \multirow{2}{*}{Demi-C\&W} & YOLOv3 & 197 & 86 & 88 & 25.0\% \\
                          & & Faster R-CNN & 337 & 215 & 187 & {\bf 47.8\%} \\
                          \cmidrule{2-7}
  & \multirow{2}{*}{Demi-MI} & YOLOv3 & 777 & 201 & 17 & {\bf 89.8\%} \\
                          & & Faster R-CNN & 776 & 111 & 9 & 82.7\% \\ \midrule

  \multirow{4}{*}[-4pt]{VGG-19}
  & \multirow{2}{*}{Demi-C\&W} & YOLOv3 & 413 & 557 & 140 & {\bf 78.8\%} \\
                          & & Faster R-CNN & 579 & 395 & 107 & 78.2\% \\
                          \cmidrule{2-7}
  & \multirow{2}{*}{Demi-MI} & YOLOv3 & 723 & 282 & 21 & {\bf 89.5\%} \\
                          & & Faster R-CNN & 746 & 182 & 13 & 82.5\% \\ \midrule

  \multirow{4}{*}[-4pt]{Inception-v3}
  & \multirow{2}{*}{Demi-C\&W} & YOLOv3 & 404 & 557 & 144 & {\bf 78.2\%} \\
                             & & Faster R-CNN & 568 & 400 & 101 & {\bf 78.2\%} \\
                             \cmidrule{2-7}
  & \multirow{2}{*}{Demi-MI} & YOLOv3 & 766 & 209 & 18 & {\bf 89.1\%}\\
                          & & Faster R-CNN & 783 & 125 & 7 & 82.4\% \\ \midrule

  \multirow{4}{*}[-4pt]{MobileNet-v2}
  & \multirow{2}{*}{Demi-C\&W} & YOLOv3 & 401 & 561 & 144 & 78.0\%\\
                             & & Faster R-CNN & 570 & 409 & 106 & {\bf 78.3\%} \\
                             \cmidrule{2-7}
  & \multirow{2}{*}{Demi-MI} & YOLOv3 & 759 & 218 & 17 & {\bf 89.1\%}\\
                          & & Faster R-CNN & 762 & 126 & 17 & 82.6\% \\
  \bottomrule
  \end{tabular}}
  \caption{We investigate the cross-task transferability of our Demiguise Attack on 1000 samples. The number of successful cases is calculated by counting the unique cases of which a sample is either mis-labeled or not labeled at all (label vanish) by the detector.}%
  \label{tab:cross-task-transferability}
\end{table}

Finally, we testify the cross-task transferability of Demiguise Attack. We use Demiguise-\{C\&W, MI-FGSM\} crafted examples on the same aforementioned classifiers and transfer them to black-box object detectors. Specifically, we use PyTorch's official Faster R-CNN~\cite{Ren2015FasterRT} (with a backbone of ResNet-50) and weights pre-trained on COCO, and we use the original weights for YOLOv3~\cite{Redmon2018YOLOv3AI} as well. We total the number of cases where a sample is (1) not labeled at all (label vanished), (2) mis-labeled, and (3) detected with extra objects. We consider the former two scenarios as successful attacks. We sum up the total number of unique samples falling in the former two cases as successful cases.

As in Table~\ref{tab:cross-task-transferability}, both Demiguise-\{C\&W, MI-FGSM\} crafted adversarial examples on all four target classifiers are able to fail detectors to a certain extent under cross-task black-box scenarios. We find that Demiguise-MI-FGSM can craft more transferable perturbations: of the 1000 samples, up to nearly 90\% are successful, which is over 100 more than Demiguise-C\&W on average. We also argue that attacking white-box classifiers and transferring adversarial examples to black-box detectors with the same backbone networks don't necessarily contribute to more transferability: Demiguise-C\&W attacking ResNet-50 don't generate as many samples that can fail detectors as the other three classifiers, even though they share the same backbone network. Nevertheless, we demonstrate that Demiguise Attack is able to achieve outstanding transferability even under cross-task black-box scenarios.

\subsection{Competition}


Proposed method ranked third in the Security AI Challenger VI Track II: Unrestricted Adversarial Attacks on ImageNet in the CVPR2021~\cite{alibaba2021}. We implement our approach for attacking an ensemble of EfficientNet~\cite{tan2019efficientnet}, ResNet-50~\cite{he2016deep}, VGG-16~\cite{simonyan2014very} and ViT~\cite{dosovitskiy2020image}. And we adopt the ensemble in the logits.

\section{Discussion}

We mentioned in Section~\ref{sec:introduction} that our approach crafts perturbations with rich semantic information, and can somewhat simulate illumination changes happening in real-world scenarios. We would like to further discuss this notion. Demiguise Attack's adversarial perturbations are optimised against Perceptual Similarity. Perceptual Similarity aligns with human perception because it utilizes deep features inside the VGG network, which learns a representation of the natural world that correlates well with perceptual judgement. Real-world objects constantly appear different because of illumination changes, and we happen to find our perturbations very similar to these. This is the primary reasoning for us to claim that our perturbations simulate some natural phenomena or photographic effects in real life. In addition, as our experiment results in Table~\ref{tab:pert_sem_rate} show, our perturbations actually contain semantic meanings to classifiers, which further implies that unknown blind spots may inherently exist inside DNN-based classifiers, and may indicate potential robustness enhancements for current widely-used DNNs.

\section{Conclusion}

In this paper, we propose a novel, unrestricted black-box adversarial attack based on Perceptual Similarity --- Demiguise Attack. Our approach can not only craft largely imperceptible perturbation by manipulating deep semantic information in high-dimensional images but also contribute to extensive enhancements when combined with existing state-of-the-art adversarial attacks. Demiguise Attack boosts adversarial strength and robustness, increases transferability for a maximum of 50\%, and shows promising cross-task transferability performances. Besides, our findings imply that semantic defenses potentially exist for DNNs, which will be discussed in our future work.

\section*{Acknowledgements}

This work is supported by the National Natural Science Foundation of China (No. 61876019, No. U1936218, and No. 62072037). We thank the security AI challenger program launched by Alibaba Group and Tsinghua University.

\bibliographystyle{named}
\bibliography{ijcai21}

\begin{thebibliography}{}

\bibitem[\protect\citeauthoryear{Carlini and Wagner}{2017}]{carlini2017towards}
Nicholas Carlini and David Wagner.
\newblock Towards evaluating the robustness of neural networks.
\newblock In {\em 2017 ieee symposium on security and privacy (sp)}, pages
  39--57. IEEE, 2017.

\bibitem[\protect\citeauthoryear{Chen \bgroup \em et al.\egroup
  }{2020}]{chen2020hopskipjumpattack}
Jianbo Chen, Michael~I Jordan, and Martin~J Wainwright.
\newblock Hopskipjumpattack: A query-efficient decision-based attack.
\newblock In {\em 2020 ieee symposium on security and privacy (sp)}, pages
  1277--1294. IEEE, 2020.

\bibitem[\protect\citeauthoryear{Das \bgroup \em et al.\egroup
  }{2018}]{Das2018SHIELDFP}
N.~Das, Madhuri Shanbhogue, Shang-Tse Chen, Fred Hohman, S.~Li, L.~Chen,
  Michael~E. Kounavis, and Duen~Horng Chau.
\newblock Shield: Fast, practical defense and vaccination for deep learning
  using jpeg compression.
\newblock {\em Proceedings of the 24th ACM SIGKDD International Conference on
  Knowledge Discovery \& Data Mining}, 2018.

\bibitem[\protect\citeauthoryear{Deng \bgroup \em et al.\egroup
  }{2009}]{Deng2009ImageNetAL}
Jia Deng, W.~Dong, R.~Socher, L.~Li, K.~Li, and Li~Fei-Fei.
\newblock Imagenet: A large-scale hierarchical image database.
\newblock In {\em CVPR 2009}, 2009.

\bibitem[\protect\citeauthoryear{Dong \bgroup \em et al.\egroup
  }{2018}]{Dong2018BoostingAA}
Y.~Dong, Fangzhou Liao, Tianyu Pang, H.~Su, J.~Zhu, Xiaolin Hu, and J.~Li.
\newblock Boosting adversarial attacks with momentum.
\newblock {\em 2018 IEEE/CVF Conference on Computer Vision and Pattern
  Recognition}, pages 9185--9193, 2018.

\bibitem[\protect\citeauthoryear{Dong \bgroup \em et al.\egroup
  }{2021}]{alibaba2021}
Y.~Dong, Q.~Fu, and X.~Yang.
\newblock Alibaba security: Adversarial robustness benchmark.
\newblock \url{https://s.alibaba.com/benchmark}, 2021.
\newblock Accessed: 2021-05-16.

\bibitem[\protect\citeauthoryear{Dosovitskiy \bgroup \em et al.\egroup
  }{2020}]{dosovitskiy2020image}
Alexey Dosovitskiy, Lucas Beyer, Alexander Kolesnikov, Dirk Weissenborn,
  Xiaohua Zhai, Thomas Unterthiner, Mostafa Dehghani, Matthias Minderer, Georg
  Heigold, Sylvain Gelly, et~al.
\newblock An image is worth 16x16 words: Transformers for image recognition at
  scale.
\newblock {\em arXiv preprint arXiv:2010.11929}, 2020.

\bibitem[\protect\citeauthoryear{Goodfellow \bgroup \em et al.\egroup
  }{2015}]{Goodfellow2015ExplainingAH}
Ian~J. Goodfellow, Jonathon Shlens, and Christian Szegedy.
\newblock Explaining and harnessing adversarial examples.
\newblock {\em CoRR}, abs/1412.6572, 2015.

\bibitem[\protect\citeauthoryear{He \bgroup \em et al.\egroup
  }{2016a}]{He2016DeepRL}
Kaiming He, X.~Zhang, Shaoqing Ren, and Jian Sun.
\newblock Deep residual learning for image recognition.
\newblock {\em 2016 IEEE Conference on Computer Vision and Pattern Recognition
  (CVPR)}, pages 770--778, 2016.

\bibitem[\protect\citeauthoryear{He \bgroup \em et al.\egroup
  }{2016b}]{he2016deep}
Kaiming He, Xiangyu Zhang, Shaoqing Ren, and Jian Sun.
\newblock Deep residual learning for image recognition.
\newblock In {\em Proceedings of the IEEE conference on computer vision and
  pattern recognition}, pages 770--778, 2016.

\bibitem[\protect\citeauthoryear{Hosseini and
  Poovendran}{2018}]{Hosseini2018SemanticAE}
H.~Hosseini and R.~Poovendran.
\newblock Semantic adversarial examples.
\newblock {\em 2018 IEEE/CVF Conference on Computer Vision and Pattern
  Recognition Workshops (CVPRW)}, pages 1695--16955, 2018.

\bibitem[\protect\citeauthoryear{Johnson \bgroup \em et al.\egroup
  }{2016}]{Johnson2016PerceptualLF}
J.~Johnson, Alexandre Alahi, and Li~Fei-Fei.
\newblock Perceptual losses for real-time style transfer and super-resolution.
\newblock In {\em ECCV}, 2016.

\bibitem[\protect\citeauthoryear{Kurakin \bgroup \em et al.\egroup
  }{2017}]{Kurakin2017AdversarialEI}
A.~Kurakin, Ian~J. Goodfellow, and S.~Bengio.
\newblock Adversarial examples in the physical world.
\newblock {\em ArXiv}, abs/1607.02533, 2017.

\bibitem[\protect\citeauthoryear{Papernot \bgroup \em et al.\egroup
  }{2016}]{Papernot2016TheLO}
Nicolas Papernot, P.~McDaniel, S.~Jha, Matt Fredrikson, Z.~Y. Celik, and
  A.~Swami.
\newblock The limitations of deep learning in adversarial settings.
\newblock {\em 2016 IEEE European Symposium on Security and Privacy
  (EuroS\&P)}, pages 372--387, 2016.

\bibitem[\protect\citeauthoryear{Redmon and Farhadi}{2018}]{Redmon2018YOLOv3AI}
Joseph Redmon and Ali Farhadi.
\newblock Yolov3: An incremental improvement.
\newblock {\em ArXiv}, abs/1804.02767, 2018.

\bibitem[\protect\citeauthoryear{Ren \bgroup \em et al.\egroup
  }{2015}]{Ren2015FasterRT}
Shaoqing Ren, Kaiming He, Ross~B. Girshick, and J.~Sun.
\newblock Faster r-cnn: Towards real-time object detection with region proposal
  networks.
\newblock {\em IEEE Transactions on Pattern Analysis and Machine Intelligence},
  39:1137--1149, 2015.

\bibitem[\protect\citeauthoryear{Sandler \bgroup \em et al.\egroup
  }{2018}]{Sandler2018MobileNetV2IR}
Mark Sandler, A.~Howard, Menglong Zhu, A.~Zhmoginov, and Liang-Chieh Chen.
\newblock Mobilenetv2: Inverted residuals and linear bottlenecks.
\newblock {\em 2018 IEEE/CVF Conference on Computer Vision and Pattern
  Recognition}, pages 4510--4520, 2018.

\bibitem[\protect\citeauthoryear{Shamsabadi \bgroup \em et al.\egroup
  }{2020}]{Shamsabadi2020ColorFoolSA}
Ali~Shahin Shamsabadi, Ricardo Sanchez-Matilla, and A.~Cavallaro.
\newblock Colorfool: Semantic adversarial colorization.
\newblock {\em 2020 IEEE/CVF Conference on Computer Vision and Pattern
  Recognition (CVPR)}, pages 1148--1157, 2020.

\bibitem[\protect\citeauthoryear{Simonyan and
  Zisserman}{2014}]{simonyan2014very}
Karen Simonyan and Andrew Zisserman.
\newblock Very deep convolutional networks for large-scale image recognition.
\newblock {\em arXiv preprint arXiv:1409.1556}, 2014.

\bibitem[\protect\citeauthoryear{Simonyan and
  Zisserman}{2015}]{Simonyan2015VeryDC}
K.~Simonyan and Andrew Zisserman.
\newblock Very deep convolutional networks for large-scale image recognition.
\newblock {\em CoRR}, abs/1409.1556, 2015.

\bibitem[\protect\citeauthoryear{Szegedy \bgroup \em et al.\egroup
  }{2014}]{Szegedy2014IntriguingPO}
Christian Szegedy, W.~Zaremba, Ilya Sutskever, Joan Bruna, D.~Erhan, Ian~J.
  Goodfellow, and R.~Fergus.
\newblock Intriguing properties of neural networks.
\newblock {\em CoRR}, abs/1312.6199, 2014.

\bibitem[\protect\citeauthoryear{Szegedy \bgroup \em et al.\egroup
  }{2016}]{Szegedy2016RethinkingTI}
Christian Szegedy, V.~Vanhoucke, S.~Ioffe, Jon Shlens, and Z.~Wojna.
\newblock Rethinking the inception architecture for computer vision.
\newblock {\em 2016 IEEE Conference on Computer Vision and Pattern Recognition
  (CVPR)}, pages 2818--2826, 2016.

\bibitem[\protect\citeauthoryear{Tan and Le}{2019}]{tan2019efficientnet}
Mingxing Tan and Quoc Le.
\newblock Efficientnet: Rethinking model scaling for convolutional neural
  networks.
\newblock In {\em International Conference on Machine Learning}, pages
  6105--6114. PMLR, 2019.

\bibitem[\protect\citeauthoryear{Wang \bgroup \em et al.\egroup
  }{2004}]{Wang2004ImageQA}
Zhou Wang, A.~Bovik, H.~R. Sheikh, and E.~P. Simoncelli.
\newblock Image quality assessment: from error visibility to structural
  similarity.
\newblock {\em IEEE Transactions on Image Processing}, 13:600--612, 2004.

\bibitem[\protect\citeauthoryear{Xie \bgroup \em et al.\egroup
  }{2019}]{Xie2019ImprovingTO}
Cihang Xie, Zhishuai Zhang, Jianyu Wang, Yuyin Zhou, Zhou Ren, and A.~Yuille.
\newblock Improving transferability of adversarial examples with input
  diversity.
\newblock {\em 2019 IEEE/CVF Conference on Computer Vision and Pattern
  Recognition (CVPR)}, pages 2725--2734, 2019.

\bibitem[\protect\citeauthoryear{Xu \bgroup \em et al.\egroup
  }{2018}]{Xu2018FeatureSD}
Weilin Xu, David Evans, and Y.~Qi.
\newblock Feature squeezing: Detecting adversarial examples in deep neural
  networks.
\newblock {\em ArXiv}, abs/1704.01155, 2018.

\bibitem[\protect\citeauthoryear{Zhang \bgroup \em et al.\egroup
  }{2011}]{Zhang2011FSIMAF}
L.~Zhang, Lei Zhang, X.~Mou, and D.~Zhang.
\newblock Fsim: A feature similarity index for image quality assessment.
\newblock {\em IEEE Transactions on Image Processing}, 20:2378--2386, 2011.

\bibitem[\protect\citeauthoryear{Zhang \bgroup \em et al.\egroup
  }{2018}]{zhang2018unreasonable}
Richard Zhang, Phillip Isola, Alexei~A Efros, Eli Shechtman, and Oliver Wang.
\newblock The unreasonable effectiveness of deep features as a perceptual
  metric.
\newblock In {\em Proceedings of the IEEE conference on computer vision and
  pattern recognition}, pages 586--595, 2018.

\end{thebibliography}

\end{document}